\documentclass[10pt,twocolumn]{IEEEtran}

\usepackage{cite}
\usepackage{amsmath,amssymb,amsfonts}
\usepackage{algorithmic}
\usepackage{graphicx}
\usepackage{textcomp}

\usepackage{makecell}
\usepackage{algorithm}
\usepackage{booktabs}
\usepackage{multirow}

\begin{document}
\title{Curvilinear Structure-preserving Unpaired Cross-domain Medical Image Translation}
\author{Zihao Chen, Yi Zhou, Xudong Jiang, Li Chen, Leopold Schmetterer,  Bingyao Tan, Jun Cheng
\thanks{Zihao Chen is with the School of Electrical and Electronic Engineering, Nanyang Technological University, Singapore, and Singapore Eye Research Institute, Singapore National Eye Centre, Singapore (email: zchen088@e.ntu.edu.sg).}
\thanks{Yi Zhou is with Singapore Eye Research Institute, Singapore National Eye Centre, Singapore (email: zhouyi.zura@gmail.com).}
\thanks{Xudong Jiang is with the School of Electrical and Electronic Engineering, Nanyang Technological University, Singapore (email: exdjiang@ntu.edu.sg).}
\thanks{Li Chen is with School of Computer Science and Technology, Wuhan University of Science and Technology, Wuhan, China (email: chenli@wust.edu.cn).}
\thanks{Leopold Schmetterer is with Singapore Eye Research Institute, Singapore National Eye Centre, Singapore; Ophthalmology \& Visual Sciences Academic Clinical Program, Duke-NUS Medical School, Singapore; Institute of Molecular and Clinical Ophthalmology, Basel, Switzerland; School of Chemical and Biomedical Engineering, Nanyang Technological University, Singapore; Centre for Medical Physics and Biomedical Engineering, Medical University of Vienna, Vienna, Austria; Department of Clinical Pharmacology, Medical University of Vienna, Vienna, Austria; and Rothschild Foundation Hospital, Paris, France (email: leopold.schmetterer@meduniwien.ac.at).}
\thanks{Bingyao Tan is with Singapore Eye Research Institute, Singapore National Eye Centre, Singapore, and Ophthalmology \& Visual Sciences Academic Clinical Program, Duke-NUS Medical School, Singapore (email: tan.bingyao@seri.com.sg).}
\thanks{Jun Cheng is with Institute for Infocomm Research (I$^2$R), A*STAR, Singapore (email: cheng\_jun@a-star.edu.sg).}
}

\maketitle

\begin{abstract}
Unpaired image-to-image translation has emerged as a crucial technique in medical imaging, enabling cross-modality synthesis, domain adaptation, and data augmentation without costly paired datasets. Yet, existing approaches often distort fine curvilinear structures, such as microvasculature, undermining both diagnostic reliability and quantitative analysis. This limitation is consequential in ophthalmic and vascular imaging, where subtle morphological changes carry significant clinical meaning.
We propose Curvilinear Structure-preserving Translation (CST), a general framework that explicitly preserves fine curvilinear structures during unpaired translation by integrating structure consistency into the training.
Specifically, CST augments baseline models with a curvilinear extraction module for topological supervision. It can be seamlessly incorporated into existing methods. We integrate it into CycleGAN and UNSB as two representative backbones. Comprehensive evaluation across three imaging modalities: optical coherence tomography angiography, color fundus and X-ray coronary angiography demonstrates that CST improves translation fidelity and achieves state-of-the-art performance. By reinforcing geometric integrity in learned mappings, CST establishes a principled pathway toward curvilinear structure-aware cross-domain translation in medical imaging.
\end{abstract}


\section{Introduction}
\label{sec:introduction}
\IEEEPARstart{C}{ross-domain} image translation has emerged as a practical solution to mitigate performance degradation caused by domain gaps between training and test data \cite{Hoffman_cycada2017,pizzati2020domain,Zhang_Tianyang,deng2024unsupervised}. In medical imaging, these gaps often arise from different scanners, imaging protocols, and imaging conditions, resulting in variations in resolution, contrast, and noise. Such discrepancies affect the generalizability of deep learning models \cite{AADD, 9113488, Liu_Yin_Qu_Wang_2023}. For example, the results of optical coherence tomography angiography (OCTA) obtained from different scanners are often incomparable, which weakens the clinical usability and generalizability of OCTA~\cite{tan2020approaches,sampson2022towards}. Furthermore, cross-domain discrepancies frequently distort or misalign fine structural details—features that are crucial for accurate diagnosis, quantitative measurement, and morphological integrity assessment. Variations in shape, appearance, and topology across datasets can substantially hinder the performance of deep learning models and reduce the reliability of inter-study comparisons \cite{li2023vesselpromoted}. Cross-domain image translation methods that specifically preserve structural consistency have shown promise in alleviating these issues, enabling more robust and transferable analysis across heterogeneous imaging modalities \cite{LI2025103404, kang2023structure,HU2024103164,Xin_Crossconditioned_MICCAI2024,chen2025mutri,KhoPoo_XOCT_MICCAI2025,ZhaXin_Structureaware_MICCAI2025}.
\begin{figure}[!t]
  \centering
  \includegraphics[width=0.50\textwidth]{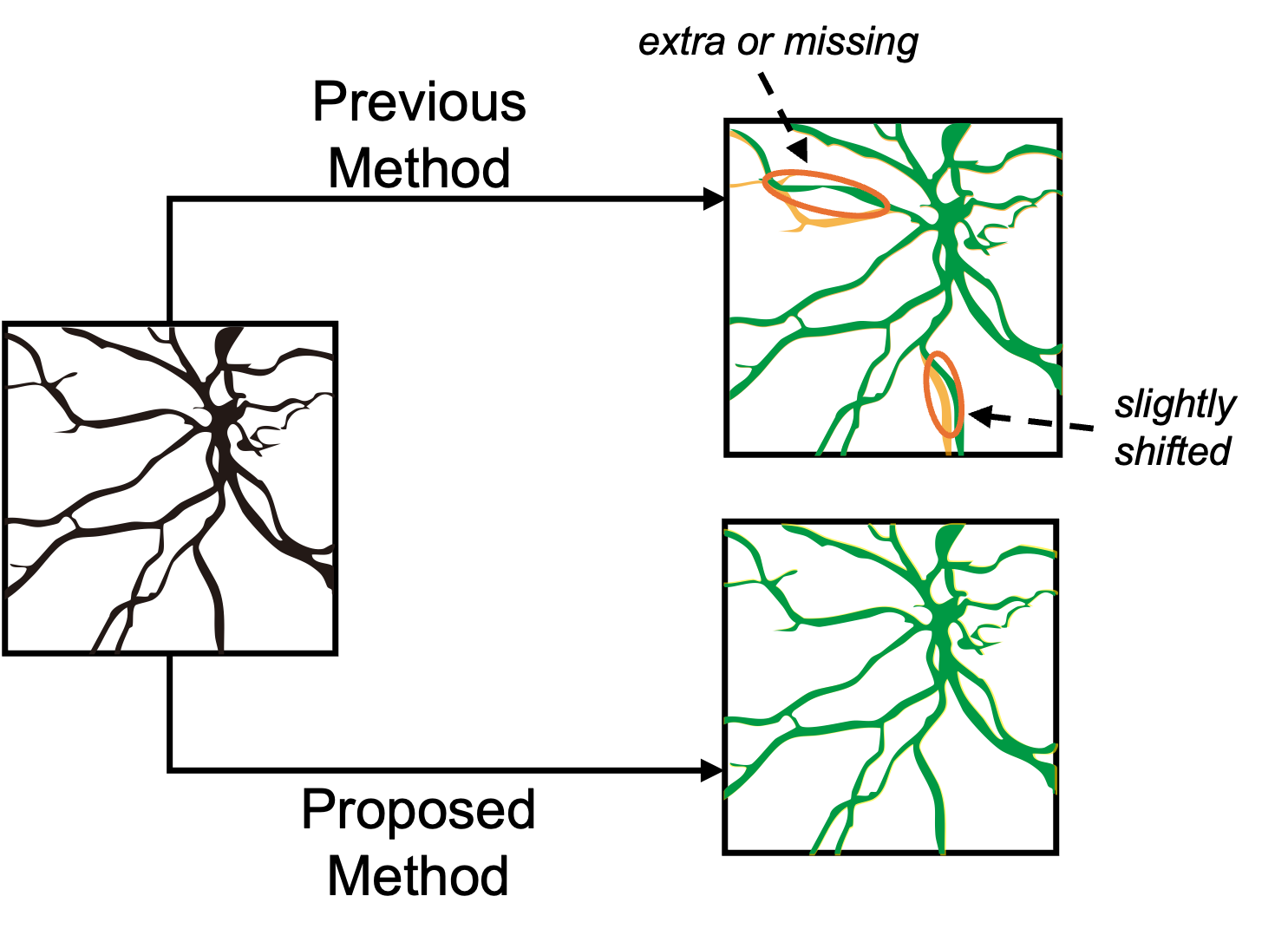}
  \caption{Illustration of the vessel fidelity issue in cross-domain translation. The black or green indicating vessels from the source and translated images respectively. Previous methods show extra/missing or shifted vessels in translated results, represented by vessels in cyan.  Our method preserves vessel structures well, making it more suitable for downstream segmentation tasks.
  }
  \label{PS}
\end{figure}

To address these challenges, image-to-image translation has been developed to learn mappings between different visual domains~\cite{pix2pix2017,upadhyay2021uncerguidedi2i,Xiao_2025_CVPR}. In some medical imaging scenarios with a scarcity of perfectly aligned datasets, paired translation methods that require pixel-wise correspondences are impractical~\cite{11141682}. Unpaired translation frameworks such as CycleGAN \cite{CycleGAN2017} and MUNIT \cite{huang2018munit} therefore emerged as attractive alternatives, enabling cross-domain mappings without paired supervision. SUANet \cite{Zhang_Tianyang} considers structure gap to account for both intensity and structural differences between the training and test sets. 
Despite their success, these methods often fail in medical data containing fine curvilinear shapes because they don't specifically encode such geometry or topology. Curvilinear structures, like vessels, nerves, and fibers, are unrepresentative because their information is high-frequency, sparse, topologically constrained, and sometimes low contrast. These networks are not inherently designed to recognize the significance of these features. As shown in Fig.~\ref{PS}, these models often synthesize vessels with distorted geometry and topology, including phantom vessels, missing vessels, and positional shifts, undermining the clinical significance of the geometric consistencies between the source and translated domains.

We propose Curvilinear Structure-preserving Translation (CST), a novel translation framework explicitly designed to maintain the integrity of curvilinear structures in unpaired medical image-to-image translation. CST innovatively imposes structural constraints during translation by leveraging features computed by a curvilinear extraction module from the pretrained universal curvilinear structure segmentation\cite{UCS} model, ensuring geometric consistency between the generated and the source images. The curvilinear extraction module  guides the image translation process, while a tailored loss function enforces structural consistency and suppresses artifacts. The key contributions of our work are summarized as follows:

\begin{itemize}
    \item We introduce a CST framework, a novel unpaired translation paradigm that explicitly preserves the curvilinear structure across domains.
    \item CST integrates a curvilinear  extraction module—adapted from the universal curvilinear structure segmentation model—into the training of representative translation frameworks such as UNSB~\cite{UNSB} and CycleGAN~\cite{CycleGAN2017}.
    \item A comprehensive loss function that combines curvilinear structure loss and rotation consistency loss is proposed for fine-grained vessel preservation.
    \item Extensive experiments across diverse imaging modalities demonstrate that CST  outperforms existing methods in preserving curvilinear structures.    
\end{itemize}

\section{Related Work}
\subsection{Unpaired Image-to-Image Translation}
Traditional image-to-image translation methods, such as Pix2Pix \cite{pix2pix2017}, rely on paired datasets with one-to-one correspondences between the source and target domains. However, collecting such pairs of data is often impractical, especially in cross-modality scenarios involving medical or industrial images, where alignment is costly or impossible. To address this, unpaired translation methods like CycleGAN \cite{CycleGAN2017}, MUNIT \cite{huang2018munit} and CUT \cite{park2020cut} learn mappings from unaligned datasets using adversarial training and domain-level constraints. Diffusion models \cite{NEURIPS2020_4c5bcfec,Song2021DenoisingDI} have recently gained attention as an alternative generative framework with strong capabilities in data distribution modeling. Compared to Generative Adversarial Networks (GANs), diffusion models are generally more stable to train and less prone to mode collapse \cite{dhariwal2021diffusionmodelsbeatgans}. While these models produce visually plausible appearance, they do not explicitly penalize changes to the shape, geometry, and connectivity of the thin structure, and subsequently they might smooth out or hallucinate vessel branches, and distort local curvature and bifurcations. These changes can alter the perceived pathology and substantially affect quantitative metrics such as fractal dimension, tortuosity, and branching angle\cite{Xin_Crossconditioned_MICCAI2024}.

Recent advances in structure-aware unpaired image translation introduced auxiliary guidance to improve spatial fidelity across domains. Specifically, UNIT-DDPM \cite{sasaki2021unitddpmunpairedimagetranslation} combines diffusion with a shared-latent space framework to align source and target domains without paired supervision. MaskGAN \cite{phan2023structure} and ContourDiff \cite{chen2024contourdiffunpairedimagetoimagetranslation} employ structural masks or contour to constrain spatial correspondence. SynDiff~\cite{ozbey_dalmaz_syndiff_2024} integrates adversarial diffusion with cycle consistency, while DDIB~\cite{su2022dual} and UNSB~\cite{UNSB} guide unpaired translation through implicit or probabilistic bridging mechanisms. Yet, these methods still fail to preserve fine curvilinear structures—such as retinal vessels in OCTA \cite{KAZEROUNI2023102846}.

\subsection{Curvilinear Structure Segmentation}

Curvilinear segmentation networks have achieved great performance in specific domains \cite{ronneberger2015unetconvolutionalnetworksbiomedical,mou2019cs,Xu_2022,LIN2023102937,zhouyi}. However, these methods lack generalization ability across modalities because they learn domain-specific feature representations that do not transfer well when appearance, contrast, or noise characteristics differ significantly between imaging modalities. Recently, foundation segmentation models have emerged, offering strong cross-domain generalization with minimal supervision. The Segment Anything Model (SAM) \cite{SAM} exemplifies this capability through its promptable, zero-shot segmentation framework. For medical images, MedSAM \cite{MedSAM} and SAM-Med2D \cite{cheng2023sammed2d} finetune SAM on specific medical image domains. Nevertheless, SAM-based methods excel at object-level segmentation but are not tailored to curvilinear structures.  More recently, the Universal Curvilinear structure Segmentation (UCS) \cite{UCS} extends from SAM to specifically target generalized segmentation of curvilinear structures, which is integrated in the current work to improve the structural fidelity during image translation. 

\begin{figure*}[!t]
  \centering
  \includegraphics[width=0.88\textwidth]{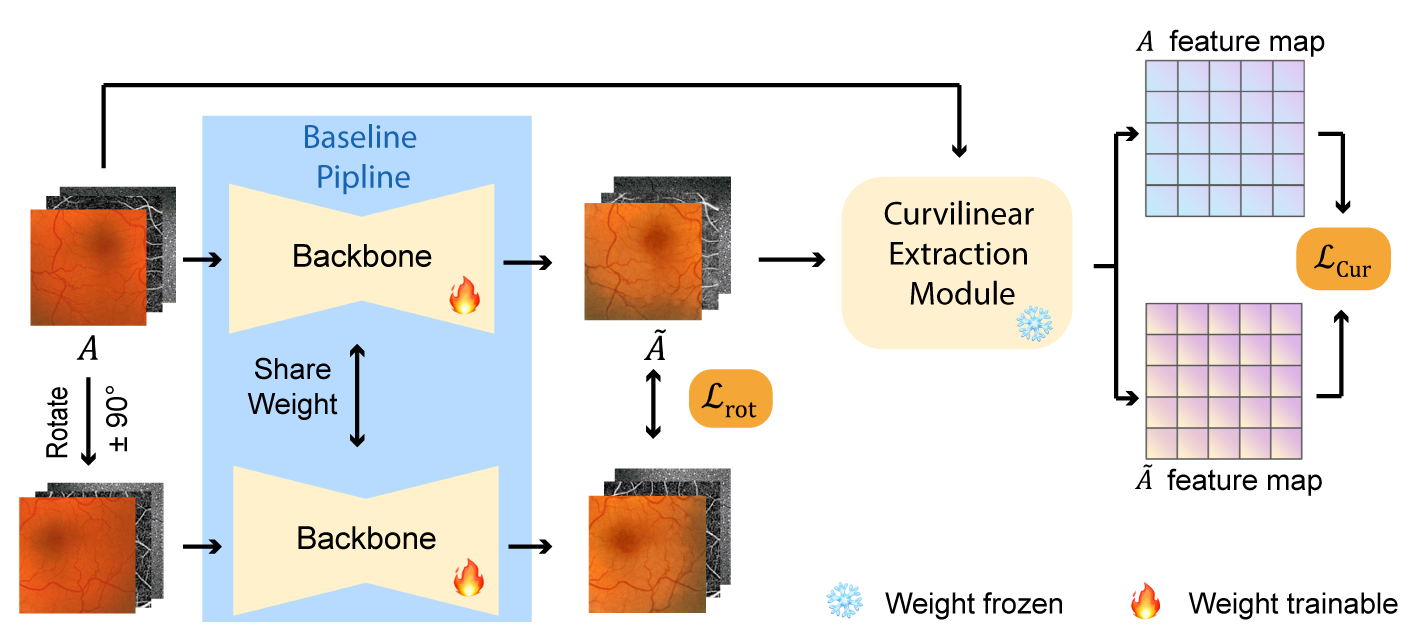}
  \caption{Overview of our proposed method. The complete training workflow. The $A$ is the image from source domain and the $\tilde{A}$ denotes the image of $A$ translated from the source domain to the target domain. The baseline pipeline illustrating the baseline model’s original forward process.
  }
  \label{model_structure}
\end{figure*}

\section{Method}
\subsection{Motivation}
The core challenge lies in the mismatch between the optimization objectives of current methods and the geometric properties of curvilinear structures. Standard generative models rely primarily on pixel-level objectives (e.g., $\ell_1$ or adversarial losses), which optimize for global visual fidelity but do not explicitly encode topological or geometric constraints. 
This creates a critical vulnerability: the generator cannot reliably distinguish between genuine vascular structures and spurious curvilinear patterns induced by imaging noise or motion artifacts. Noise from acquisition or electronic interference often produces random intensity fluctuations, while motion artifacts from patient movement or physiological processes create streaking and blurring effects that visually resemble thin vessels. Since pixel-based objectives treat all curvilinear patterns equivalently regardless of their origin, the model may misinterpret artifacts as genuine vascular information and preserve them, while simultaneously failing to maintain true vessel structures that appear similar in pixel intensity but carry anatomical significance. This results in translated images where vascular topology is corrupted, motivating the need for explicit structural guidance mechanisms.

\subsection{Architecture of Curvilinear Structure-preserving Translation}

We propose Curvilinear Structure-preserving Translation (CST), a framework that enhances existing unpaired image translation methods with structural consistency. 
CST addresses this limitation by augmenting the baseline training objectives with two 
complementary regularization terms that enforce structural, geometric, and perceptual 
consistency. As illustrated in Fig.~\ref{model_structure}, when CST is integrated with a baseline 
model, the translation process is enhanced with structure-aware guidance through two key components. 
First, a curvilinear extraction module  explicitly captures topological information from both input and generated images, preserving vascular topology through structural supervision with curvilinear structure loss.
Second, a rotation consistency flow enforces geometric invariance by requiring the model to produce consistent outputs under orthogonal transformations, preventing orientation-dependent artifacts.

\subsubsection{Curvilinear Extraction Module}
\begin{figure}[!t]
  \centering
  \includegraphics[width=0.45\textwidth]{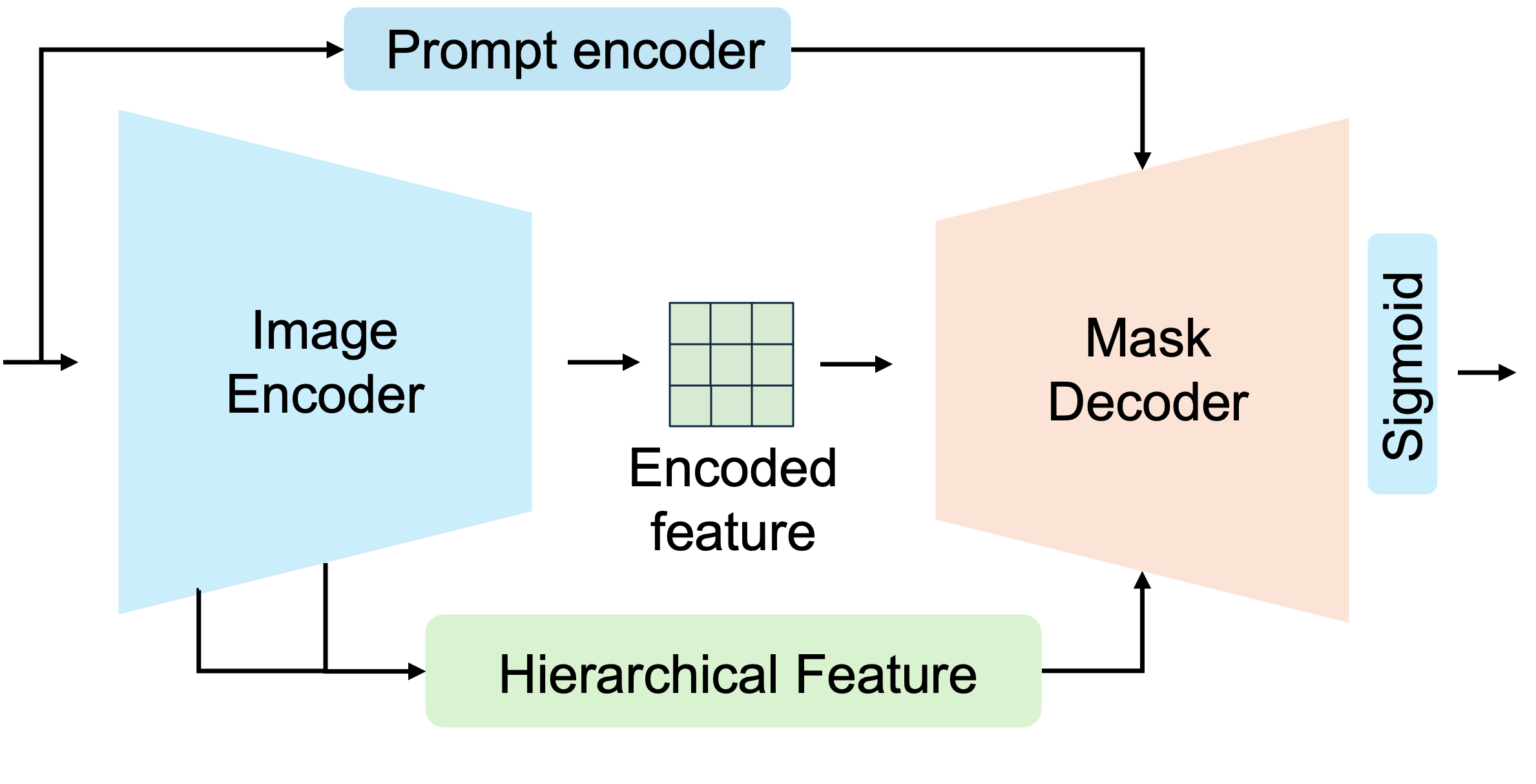}
  \caption{Architecture of the Curvilinear Extraction Module (CEM). We utilize the mask decoder output with sigmoid activation as CEM, which captures curvilinear structural semantics rather than low-level textural patterns from intermediate features.
  }
  \label{CEM}
\end{figure}

To address the fundamental limitations of pixel-level optimization in preserving curvilinear structures, we introduce an explicit structural guidance mechanism. This ambiguity manifests in two critical failure modes: (i) false preservation of noise-induced curvilinear artifacts that mimic vessel appearance, and (ii) unintended degradation of authentic anatomical structures during translation. To mitigate these issues, we inject explicit topological and geometric priors into the generative network through a dedicated Curvilinear Extraction Module (CEM).
Given the recent success of foundation models in computer vision, we initially experimented with the Segment Anything Model (SAM)~\cite{SAM} and SAM-Med2D~\cite{cheng2023sammed2d} to construct CEM.  However, preliminary experiments revealed that despite their strong generalization capabilities in  segmentation, it struggles to preserve curvilinear structures. Upon further analysis, we found that while SAM and SAM-Med2D perform well on many images, they are not optimized for domains characterized by fine, elongated structures. To address this limitation, we leverage recent advancements in curvilinear structure segmentation and adopt the Universal Curvilinear structure Segmentation (UCS) model~\cite{UCS} for robust, general-purpose extraction of curvilinear features. UCS is specifically designed and trained to capture the geometric and topological characteristics of curvilinear patterns across diverse imaging domains. This specialization is crucial for our task: UCS's cross-domain pre-training on various curvilinear structures enables it to learn domain-agnostic geometric properties that are invariant to imaging modality and noise characteristics. 
We have also explored using different stages of the UCS to construct the CEM. 
The output of the mask decoder from UCS is adopted as it contains information from both the hierarchical features and prompt-guided embeddings to produce semantically rich structural information that is specifically attuned to curvilinear geometries. Consequently, we apply a sigmoid activation to normalize it into probabilistic structural maps to compute a structure-preserving loss. Fig.~\ref{CEM} illustrates the overall structure of CEM. 
Given an input image $\mathbf{x} \in \mathbb{R}^{H \times W \times C}$, CEM produces a probability map $\mathbf{M}_{\text{curv}} \in [0,1]^{H \times W}$ highlighting prominent curvilinear structures:
\begin{equation}
\mathbf{M}_{\text{curv}} = \text{CEM}(\mathbf{x}),
\label{eq:ucs_extraction}
\end{equation}
where high probability values indicate regions exhibiting strong curvilinear geometric coherence. 

Critically, this structural map encodes topological information that remains invariant to imaging modality, acquisition protocols, and noise characteristics, effectively filtering spurious artifact patterns while preserving geometrically consistent structures.

\subsubsection{Rotation consistency flow}
Curvilinear structures in medical images, such as vessels, should remain invariant under spatial transformations. However, standard methods often fail to maintain this invariance, manifesting as preferential degradation of vessels aligned with certain orientations or generation of angle-dependent artifacts. To address this, we introduce a rotation consistency flow that enforces equivariance under orthogonal rotations, ensuring the generator produces geometrically consistent outputs regardless of structural orientation. Rotation augmentation is a well-established technique in classification and detection tasks. 
Given an input image $\mathbf{x} \in \mathbb{R}^{H \times W \times C}$, we randomly rotate it $\mathcal{R} \in \{90^\circ \times n \mid n \in \{1, 2, 3\}\}$ into $\mathcal{R}(\mathbf{x})$. We use the rotation to multiples of $90^\circ$ instead of random angles to avoid associated interpolation artifacts and boundary padding issues. 
The original and rotated images are processed independently through the generator $G(\cdot)$, which performs a complete generative mapping from the source domain to the target domain. The output from the rotated branch is subsequently restored to the canonical orientation by applying the inverse rotation  $\mathcal{R}^{-1}$:
\begin{equation}
\hat{\mathbf{I}}_{\text{ori}} = G(\mathbf{x}), \quad \hat{\mathbf{I}}_{\text{rot}} = \mathcal{R}^{-1}(G(\mathcal{R}(\mathbf{x}))),
\label{eq:rotation_flow}
\end{equation}

\subsection{Loss Functions}

We propose two loss functions tailored to guide the model toward generating realistic images while preserving essential structural information.

\subsubsection{Curvilinear structure loss}
Given an input image and a generated image, we employ a dual-level structural preservation strategy that captures both fine-grained curvilinear patterns and holistic perceptual features. We use the CEM to extract the corresponding structural probability maps $M_{\text{r}}$ and $M_{\text{f}}$ from the input and the generated image respectively. To encourage perceptual fidelity and enhance the semantic realism of the generated images, we propose to compute a curvilinear structure loss by combining the similarities between the feature maps and the Learned Perceptual Image Patch Similarity (LPIPS) loss \cite{zhang2018perceptual} between the input and output images.
Specifically, we define the curvilinear structure loss as:

\begin{equation}
\mathcal{L}_{\text{Cur}} = \mathcal{L}_{\text{dice}}(M_{\text{f}}, M_{\text{r}}) + \mathcal{L}_{\text{iou}}(M_{\text{f}}, M_{\text{r}}) + \text{LPIPS}(I_{\text{r}}, I_{\text{f}}),
\end{equation}
where $I_{\text{r}}$ denote the input image and $I_{\text{f}}$ denote the generated output. The Dice loss and the IoU loss are formulated as:
\begin{equation}
\mathcal{L}_{\text{dice}} = 1 - \frac{2 \sum_i M_i^{\text{f}} M_i^{\text{r}} + \epsilon}{\sum_i M_i^{\text{f}} + \sum_i M_i^{\text{r}} + \epsilon},
\end{equation}

\begin{equation}
\mathcal{L}_{\text{iou}} = 1 - \frac{\sum_i M_i^{\text{f}} M_i^{\text{r}} + \epsilon}{\sum_i M_i^{\text{f}} + \sum_i M_i^{\text{r}} - \sum_i M_i^{\text{f}} M_i^{\text{r}} + \epsilon},
\end{equation}
where $\epsilon$ is a small constant added for numerical stability.

This combined loss function enforces structural consistency at two complementary levels. At the curvilinear detail level, the Dice loss and IoU loss computed on UCS-derived probability maps ensure precise preservation of fine-grained curvilinear structures such as blood vessels. At the holistic structural level, the LPIPS loss~\cite{zhang2018perceptual} computed between the input image $I_{\text{r}}$ and generated image $I_{\text{f}}$ captures global structural similarity through deep features from a pretrained VGG network~\cite{simonyan2015vgg}, measuring perceptual coherence in terms of spatial arrangement and overall structure. By constraining the model on both fine curvilinear details and global perceptual structure, this dual-level loss provides more stable and comprehensive guidance during training, ensuring that the generated images preserve both curvilinear structural accuracy and perceptual realism.

\subsubsection{Rotation consistency loss}
The rotation consistency loss is defined as the $\ell_1$ distance between the two aligned outputs:

\begin{equation}
\mathcal{L}_{\text{rot}} = \| \hat{I}_{\text{ori}} - \hat{I}_{\text{rot}} \|_1.
\end{equation}

This loss encourages the model to learn a rotation-consistent mapping, effectively regularizing the output space and promoting structural invariance under rigid transformations.

\begin{table*}[t]
\caption{Structural metrics comparison on OCTA500 $\leftrightarrow$ ROSE-1, DRIVE $\leftrightarrow$ STARE, and DCA1 $\leftrightarrow$ XCAD translation tasks. Bold indicates the best performance, while the second-best result with an underline in each column.}
\centering
\scriptsize
\setlength{\tabcolsep}{2mm}
\begin{tabular}{l|cc|cc|cc|cc|cc|cc}
\toprule
\multirow{2}{*}{\textbf{Method}} & 
\multicolumn{2}{c|}{OCTA500 $\rightarrow$ ROSE-1} &
\multicolumn{2}{c|}{ROSE-1 $\rightarrow$ OCTA500} &
\multicolumn{2}{c|}{DRIVE $\rightarrow$ STARE} &
\multicolumn{2}{c|}{STARE $\rightarrow$ DRIVE} &
\multicolumn{2}{c|}{DCA1 $\rightarrow$ XCAD} &
\multicolumn{2}{c}{XCAD $\rightarrow$ DCA1} \\
\cmidrule(lr){2-3}
\cmidrule(lr){4-5}
\cmidrule(lr){6-7}
\cmidrule(lr){8-9}
\cmidrule(lr){10-11}
\cmidrule(lr){12-13}
& $\downarrow$LPIPS & $\uparrow$SSIM 
& $\downarrow$LPIPS & $\uparrow$SSIM 
& $\downarrow$LPIPS & $\uparrow$SSIM 
& $\downarrow$LPIPS & $\uparrow$SSIM 
& $\downarrow$LPIPS & $\uparrow$SSIM 
& $\downarrow$LPIPS & $\uparrow$SSIM \\
\midrule
CycleGAN \cite{CycleGAN2017}      
& 0.189 & 0.538 & 0.218 & \underline{0.578} & 0.256 & 0.644 & 0.245 & 0.665 
& 0.305 & 0.645 & 0.270 & 0.665 \\
CUT \cite{park2020cut}         
& 0.208 & 0.518 & 0.219 & 0.521 & 0.215 & 0.631 & 0.273 & 0.668 
& 0.357 & 0.652 & 0.273 & 0.683 \\
SynDiff \cite{ozbey_dalmaz_syndiff_2024}      
& 0.264 & 0.076 & 0.160 & 0.527 & 0.395 & 0.665 & 0.282 & 0.703 
& 0.357 & 0.652 & 0.379 & 0.639 \\
UNSB \cite{UNSB} 
& 0.229 & 0.442 & 0.241 & 0.397 & 0.292 & 0.653 & 0.272 & 0.684 
& 0.268 & 0.686 & 0.263 & 0.683 \\
STABLE \cite{STABLE} & 0.231 & 0.505 & 0.175 & 0.544 & 0.357 & 0.660 & 0.222 & 0.675 & 0.298 & 0.683 & 0.304 & 0.718 \\
\midrule
\textbf{Ours(UNSB)} 
& \underline{0.145} & \underline{0.540} & \underline{0.144} & 0.477 & \textbf{0.168} & \textbf{0.807} & \underline{0.206} & \textbf{0.733} 
& \underline{0.246} & \underline{0.721} & \underline{0.209} & \underline{0.746} \\
\textbf{Ours(CycleGAN)} 
& \textbf{0.076} & \textbf{0.805} & \textbf{0.114} & \textbf{0.687} & \underline{0.183} & \underline{0.729} & \textbf{0.175} & \underline{0.728} 
& \textbf{0.215} & \textbf{0.756} & \textbf{0.183} & \textbf{0.782} \\
\bottomrule
\end{tabular}
\label{tab:quality_octadrive}
\end{table*}

\subsubsection{Full objective}
The overall training objective combines the baseline model's original loss with two additional regularization terms. Here, $\mathcal{L}_{\text{base}}$ represents the original loss function of the baseline model. The full loss function is defined as:

\begin{equation}
\mathcal{L}_{\text{total}} = \mathcal{L}_{\text{base}} + \lambda_{\text{1}} \mathcal{L}_{\text{Cur}} + \lambda_{\text{2}} \mathcal{L}_{\text{rot}}.
\end{equation}

The weights $\lambda_{\text{1}}$ and $\lambda_{\text{2}}$  are hyperparameters that balance the contribution of each regularization term. We empirically set $\lambda_{\text{1}}=1.0$ and $\lambda_{\text{2}}=1.0$ to balance the magnitudes of different loss terms such that none of the items dominates the results.

\section{Experiments}

\begin{table*}[t]
\caption{Segmentation results (mean $\pm$ std, reported in \%). The results are shown for translated images from OCTA500 $\leftrightarrow$ ROSE-1 and DRIVE $\leftrightarrow$ STARE. The segmentation model was trained on the target domain. Bold indicates the best performance, while the second-best result with an underline in each column.}
\centering
\scriptsize
\setlength{\tabcolsep}{4pt}
\begin{tabular}{l|ccc|ccc|ccc|ccc}
\toprule
\multirow{2}{*}{\textbf{Method}} & 
\multicolumn{3}{c|}{OCTA500 $\rightarrow$ ROSE-1} &
\multicolumn{3}{c|}{ROSE-1 $\rightarrow$ OCTA500} &
\multicolumn{3}{c|}{DRIVE $\rightarrow$ STARE} &
\multicolumn{3}{c}{STARE $\rightarrow$ DRIVE} \\
\cmidrule(lr){2-4}
\cmidrule(lr){5-7}
\cmidrule(lr){8-10}
\cmidrule(lr){11-13}
& $\uparrow$mDice & $\uparrow$mIoU & $\uparrow$mclDice 
& $\uparrow$mDice & $\uparrow$mIoU & $\uparrow$mclDice 
& $\uparrow$mDice & $\uparrow$mIoU & $\uparrow$mclDice 
& $\uparrow$mDice & $\uparrow$mIoU & $\uparrow$mclDice \\
\midrule
No Translation & 30.4$\pm$12.0 & 18.5$\pm$8.5 & 32.7$\pm$13.5 & 37.6$\pm$8.6 & 23.4$\pm$6.2 & 35.0$\pm$7.8 & 53.7$\pm$17.8 & 38.5$\pm$15.3 & 51.2$\pm$18.4 & 54.3$\pm$26.6 & 41.3$\pm$22.3 & 55.4$\pm$27.1 \\
CycleGAN \cite{CycleGAN2017} & 39.3$\pm$4.3 & 24.5$\pm$3.3 & 42.3$\pm$4.7 & 23.7$\pm$3.5 & 13.5$\pm$2.3 & 20.7$\pm$4.1 & 45.9$\pm$12.6 & 30.6$\pm$10.3 & 43.1$\pm$13.1 & 55.3$\pm$12.8 & 39.3$\pm$11.6 & 57.6$\pm$12.5 \\
CUT \cite{park2020cut} & 38.5$\pm$4.8 & 24.0$\pm$3.7 & 41.1$\pm$5.1 & 29.6$\pm$2.4 & 17.4$\pm$1.7 & 26.5$\pm$4.2 & 47.1$\pm$13.3 & 31.8$\pm$11.6 & 45.3$\pm$12.3 & 57.1$\pm$12.8 & 41.0$\pm$12.2 & 61.1$\pm$12.4 \\
SynDiff \cite{ozbey_dalmaz_syndiff_2024} & 22.5$\pm$3.4 & 12.7$\pm$2.2 & 21.4$\pm$3.4 & 35.2$\pm$3.9 & 21.4$\pm$2.8 & 32.1$\pm$5.2 & 29.3$\pm$12.5 & 17.8$\pm$8.6 & 24.3$\pm$11.1 & 45.9$\pm$10.5 & 30.4$\pm$8.5 & 45.8$\pm$12.8 \\
UNSB \cite{UNSB} & 39.5$\pm$5.0 & 24.7$\pm$3.9 & 42.0$\pm$5.5 & 32.7$\pm$5.4 & 19.7$\pm$3.9 & 30.2$\pm$6.6 & 50.9$\pm$10.1 & 34.7$\pm$8.4 & 50.6$\pm$9.8 & 53.2$\pm$10.7 & 36.9$\pm$9.5 & 57.0$\pm$11.2 \\
STABLE \cite{STABLE} & 42.6$\pm$11.0 & 27.6$\pm$8.1 & 46.8$\pm$12.9 & 40.8$\pm$3.3 & 25.7$\pm$2.6 & 35.8$\pm$5.4 & 33.5$\pm$11.4 & 20.7$\pm$8.3 & 31.8$\pm$11.1 & 37.2$\pm$18.6 & 24.4$\pm$13.6 & 39.8$\pm$20.1 \\
\midrule
\textbf{Ours(UNSB)} & \underline{45.1$\pm$4.4} & \underline{29.2$\pm$3.7} & \underline{49.8$\pm$5.2} & \textbf{47.1$\pm$5.9} & \textbf{31.0$\pm$5.1} & \textbf{43.3$\pm$9.8} & \textbf{62.4$\pm$6.3} & \textbf{45.6$\pm$6.5} & \textbf{60.8$\pm$7.5} & \underline{62.5$\pm$11.4} & \underline{46.4$\pm$11.4} & \underline{67.0$\pm$11.3} \\
\textbf{Ours(CycleGAN)} & \textbf{49.1$\pm$5.0} & \textbf{32.6$\pm$4.3} & \textbf{55.2$\pm$6.1} & \underline{44.9$\pm$5.3} & \underline{29.1$\pm$4.4} & \underline{40.9$\pm$8.3} & \underline{57.1$\pm$11.2} & \underline{40.7$\pm$10.4} & \underline{54.7$\pm$13.2} & \textbf{63.8$\pm$14.3} & \textbf{48.3$\pm$14.2} & \textbf{68.9$\pm$13.9} \\
\bottomrule
\end{tabular}
\label{tab:seg_octafundus}
\end{table*}

\begin{table*}[t]
\caption{Segmentation results (mean $\pm$ std, reported in \%). The results are shown for translated images from DCA1 $\leftrightarrow$ XCAD. The segmentation model was trained on the target domain. Bold indicates the best performance, while the second-best result with an underline in each column.}
\centering
\setlength{\tabcolsep}{6pt}
\begin{tabular}{l|ccc|ccc}
\toprule
\multirow{2}{*}{\textbf{Method}} & 
\multicolumn{3}{c|}{DCA1 $\rightarrow$ XCAD} &
\multicolumn{3}{c}{XCAD $\rightarrow$ DCA1} \\
\cmidrule(lr){2-4}
\cmidrule(lr){5-7}
& $\uparrow$mDice & $\uparrow$mIoU & $\uparrow$mclDice 
& $\uparrow$mDice & $\uparrow$mIoU & $\uparrow$mclDice \\
\midrule
No Translation & 40.5$\pm$25.5 & 28.6$\pm$20.3 & 46.5$\pm$28.3 & \underline{55.4$\pm$6.9} & 38.6$\pm$6.4 & \underline{62.9$\pm$8.4} \\
CycleGAN \cite{CycleGAN2017} & 46.2$\pm$10.7 & 30.6$\pm$9.0 & 51.2$\pm$10.9 & 51.9$\pm$11.1 & 35.8$\pm$9.6 & 57.4$\pm$11.6 \\
CUT \cite{park2020cut} & 34.9$\pm$13.5 & 22.0$\pm$9.9 & 37.5$\pm$13.4 & 50.0$\pm$6.7 & 33.6$\pm$6.0 & 54.6$\pm$8.4 \\
SynDiff \cite{ozbey_dalmaz_syndiff_2024} & 48.7$\pm$8.8 & 32.6$\pm$7.5 & 56.1$\pm$10.9 & 42.2$\pm$7.2 & 27.0$\pm$5.7 & 41.7$\pm$8.6 \\
UNSB \cite{UNSB} & 51.1$\pm$10.5 & 35.0$\pm$9.3 & 57.9$\pm$11.0 & 55.0$\pm$8.4 & 38.4$\pm$7.6 & 59.2$\pm$8.9 \\ 
STABLE \cite{STABLE} &  50.3$\pm$15.5 & 35.0$\pm$13.3 & 55.2$\pm$18.1 & 51.1$\pm$8.5 & 34.8$\pm$7.4 & 57.5$\pm$9.3 \\
\midrule
\textbf{Ours(UNSB)} & \underline{53.5$\pm$11.3} & \underline{37.3$\pm$10.1} & \underline{59.9$\pm$11.6} & 55.3$\pm$8.6 & \underline{38.7$\pm$8.1} & 59.8$\pm$9.0 \\
\textbf{Ours(CycleGAN)} & \textbf{60.6$\pm$9.2} & \textbf{44.1$\pm$9.0} & \textbf{67.2$\pm$10.2} & \textbf{58.5$\pm$7.6} & \textbf{41.7$\pm$7.7} & \textbf{63.4$\pm$8.0} \\
\bottomrule
\end{tabular}
\label{tab:seg_dca1xcad}
\end{table*}
\begin{figure*}[t]
  \centering
  \includegraphics[width=0.96\textwidth]{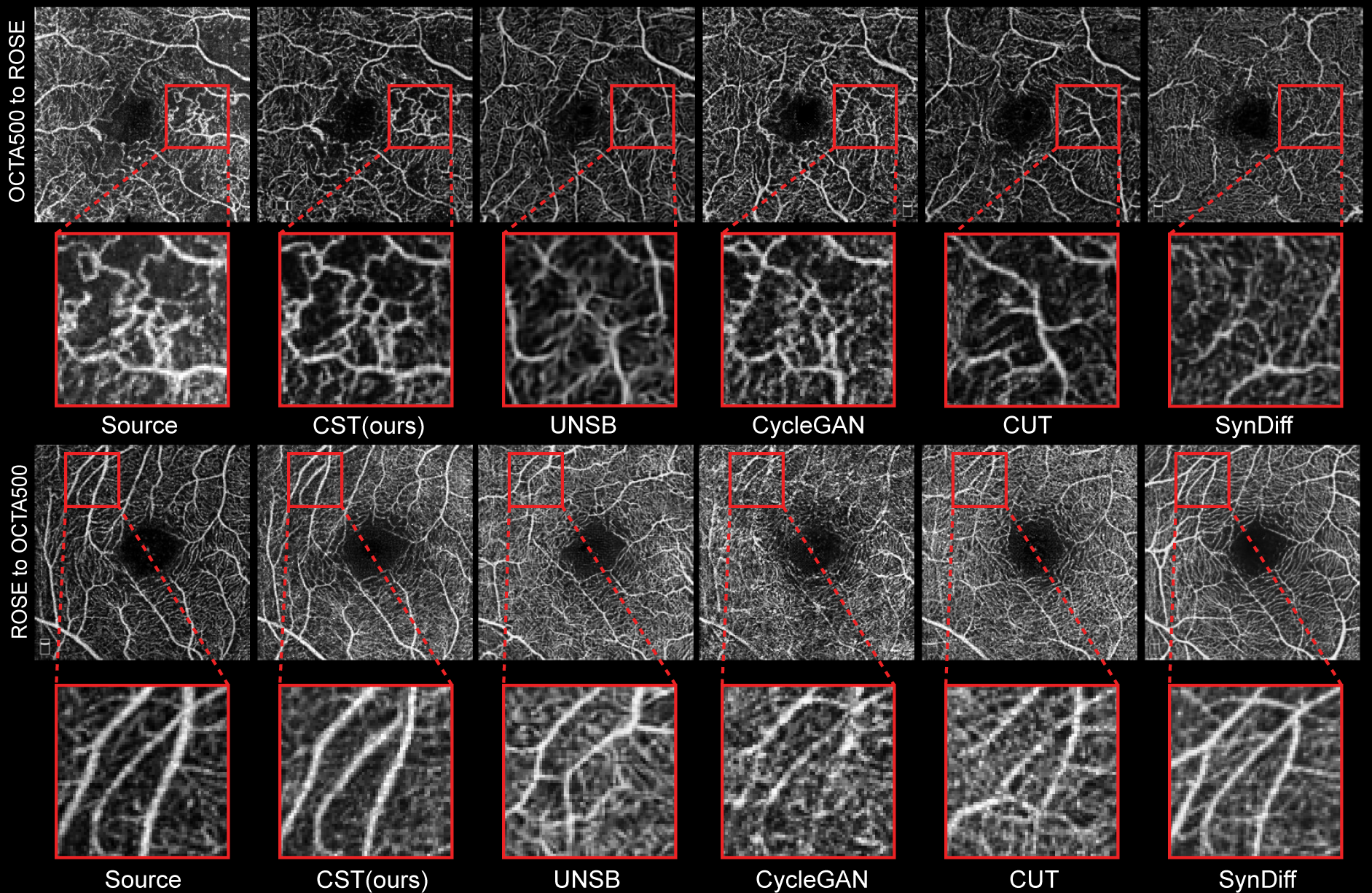}
  \caption{Qualitative results on the OCTA translation task (OCTA500$\leftrightarrow$ROSE). The OCTA500 to ROSE case corresponds to an eye with  diabetic retinopathy, while the ROSE to OCTA500 case corresponds to a healthy eye. Other methods tend to distort curvilinear structures during translation, while ours preserves them more effectively.
  }
  \label{image_results}
\end{figure*}
\begin{figure*}[t]
  \centering
  \includegraphics[width=0.98\textwidth]{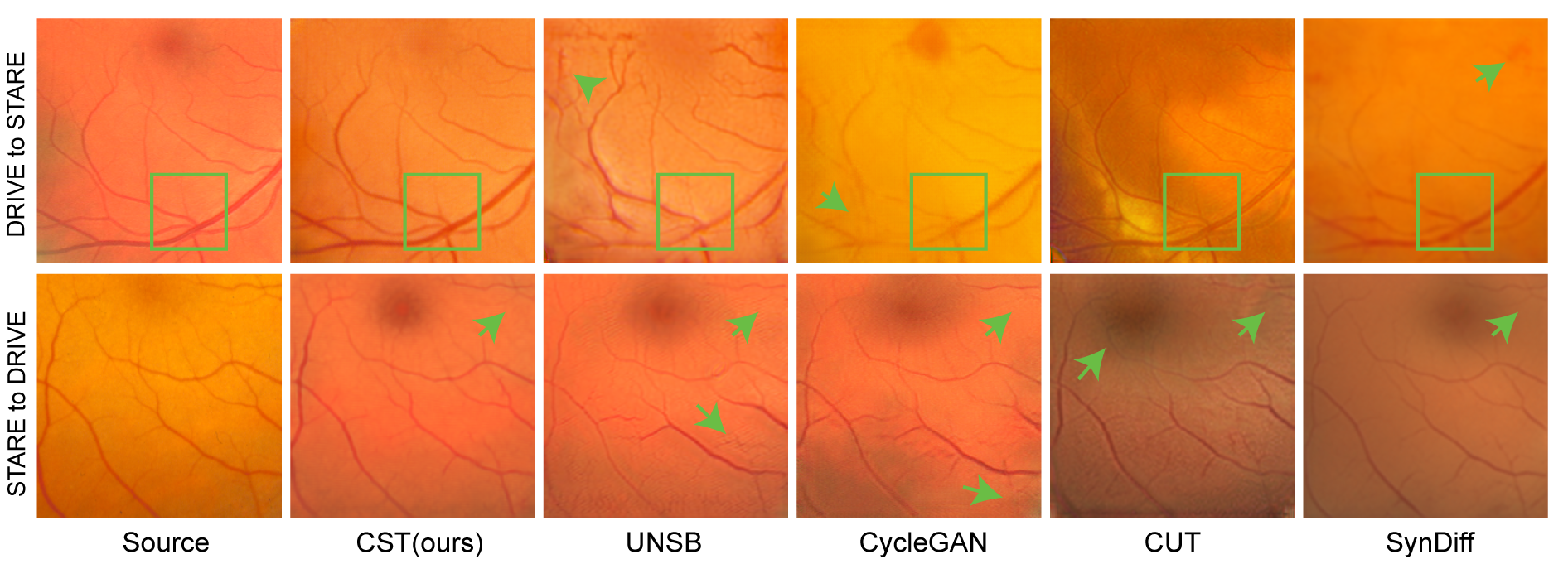}
  \caption{Qualitative results on the fundus translation task (DRIVE$\leftrightarrow$STARE). In particular, the regions highlighted by green boxes and arrowheads show that our model better maintains the original curvilinear structures from the source images.
  }
  \label{fundus_results}
\end{figure*}

\begin{figure*}[t]
  \centering
  \includegraphics[width=0.98\textwidth]{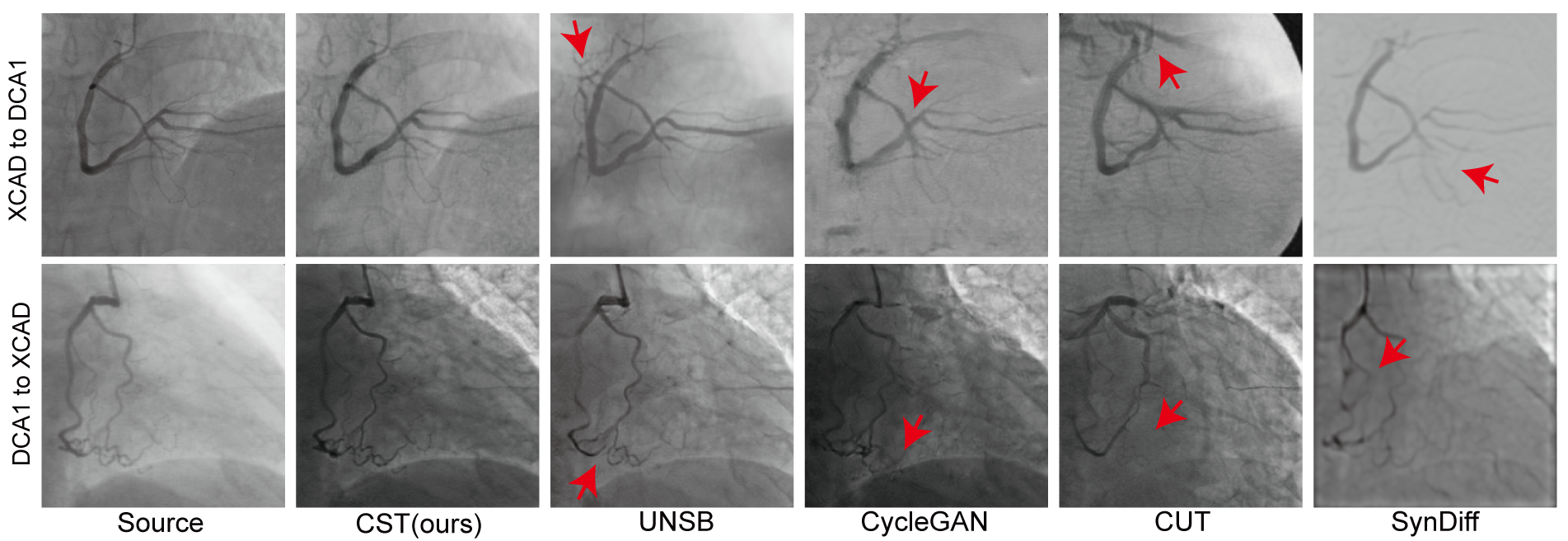}
  \caption{Qualitative results on the XCA translation task (DCA1$\leftrightarrow$XCAD). In particular, the regions highlighted by red arrowheads show that our model better maintains the original curvilinear structures from the source images.
  }
  \label{xca}
\end{figure*}

We conduct extensive experiments to evaluate the effectiveness of our proposed method in unpaired domain translation tasks. CST is validated on two representative baseline architectures: CycleGAN~\cite{CycleGAN2017}, a widely-used adversarial framework, and UNSB~\cite{UNSB}, a recent diffusion-based approach. Specifically, we compare our approach against several representative and state-of-the-art baselines, including CycleGAN~\cite{CycleGAN2017}, CUT~\cite{park2020cut}, SynDiff~\cite{ozbey_dalmaz_syndiff_2024}, UNSB~\cite{UNSB} and STABLE~\cite{STABLE}.

To assess the performance, we evaluate both image-level translation quality and task-level utility. For image-level evaluation, we report the LPIPS~\cite{zhang2018perceptual} and the Structural Similarity Index (SSIM)~\cite{SSIM}. For task-level evaluation, we adopt a downstream segmentation to measure the domain transfer quality in terms of Dice coefficient (Dice)~\cite{chen2014relationoptimaltransportschrodinger}, Intersection over Union (IoU)~\cite{MaskR-CNN} and centerline Dice (clDice)~\cite{shit2021cldice}. Our method consistently achieves good performance across all metrics and datasets, demonstrating state-of-the-art results in both perceptual fidelity and downstream segmentation effectiveness.

\subsection{Datasets and Tasks}
We evaluate our method across three distinct visual domains with structural relevance: OCTA, color fundus and X-ray Coronary angiography (XCA). Each domain comprises two publicly available datasets. We perform unpaired image translation in both directions between each dataset pair, resulting in a total of six translation tasks. The selected datasets are:

\begin{itemize}
    \item OCTA: We use OCTA-500 \cite{li2019octa} and ROSE \cite{ma2021rose:}, two OCTA datasets that capture retinal vasculature under different resolution, vascular density, and image quality.
    
    \item Color Fundus: The DRIVE \cite{DRIVE} and STARE \cite{STARE} datasets consist of retinal fundus photographs with vascular structures, collected under varying illumination and acquisition conditions. Their variation in color tone and vessel contrast makes them ideal for evaluating structure-preserving translation.

    \item XCA: The DCA1~\cite{cervantes2019automatic} and XCAD~\cite{ma2021self} datasets consist of X-ray coronary angiograms with expert-annotated vascular structures, collected under varying contrast and imaging conditions. 

\end{itemize}

During training, no paired images are used. We train the model to perform unpaired bidirectional translation between each dataset pair (e.g., OCTA-500 $\leftrightarrow$ ROSE-1), allowing evaluation of both generalization and structural consistency across different visual domains.

\subsection{Implementation Details}
All models are trained using a single NVIDIA H100 NVL GPU. For each domain (OCTA, color fundus and XCA), we train our network from scratch using unpaired images. For the OCTA and XCA dataset, all images are resized to $256 \times 256$ before being used for model training. For the OCTA, translation was conducted between the ROSE-1~\cite{ma2021rose:} and the OCTA-500~\cite{li2019octa} with a 3$\times$3~mm$^2$ field of view. In contrast, for the color fundus dataset, $256 \times 256$ patches are randomly cropped from the region of interest and then used as training inputs. The number of training iterations is set to 400 epochs for each modality. We use the Adam optimizer with $\beta_1 = 0.5$, $\beta_2 = 0.999$, and an initial learning rate of $2 \times 10^{-4}$. Following common practice, the learning rate remains constant for the first 200 epochs and is then linearly decayed to zero over the remaining training steps.
To evaluate the utility of the translated images in downstream tasks, we train a U-Net \cite{ronneberger2015unetconvolutionalnetworksbiomedical} model to segment large vessels on the target domain. This setup enables us to assess whether domain translation improves segmentation performance in cross-domain settings.

\subsection{Quantitative Evaluation}

To assess the perceptual similarity and low-level structural preservation between input and translated images, we compute the LPIPS \cite{zhang2018perceptual} and SSIM~\cite{SSIM}. As shown in Table~\ref{tab:quality_octadrive}, our method demonstrates consistent superior performance across multiple translation tasks across three modalities (OCTA, fundus and XCA). We performed t-tests comparing our method with each baseline, revealing significant differences ($p < 0.001$). Notably, compared to the best baseline methods, our approach achieves substantial improvements in LPIPS (from 0.189 to 0.076) and in SSIM (from 0.538 to 0.805). 

To evaluate the structural fidelity and perceptual quality of the translated images, we perform a comprehensive quantitative analysis encompassing both task-driven and image-level metrics. For structural consistency in downstream applications, we employ a semantic segmentation model trained on the target domain. This model is then applied directly to the translated images, and the predicted segmentation masks are quantitatively compared against ground truth annotations using the mean Dice (mDice) \cite{chen2014relationoptimaltransportschrodinger}, mean IoU (mIoU)~\cite{MaskR-CNN} and mean clDice (mclDice)~\cite{shit2021cldice}. As shown in Table~\ref{tab:seg_octafundus} and Table~\ref{tab:seg_dca1xcad}, our methods by combining CST with CycleGAN and UNSB  achieve   better performance across all tasks. We evaluated the statistical significance using image-wise t-tests between our method and the baselines, and all improvements were found to be   significant ($p<0.001$).
\begin{figure}[t]
  \centering
  \includegraphics[width=0.48\textwidth]{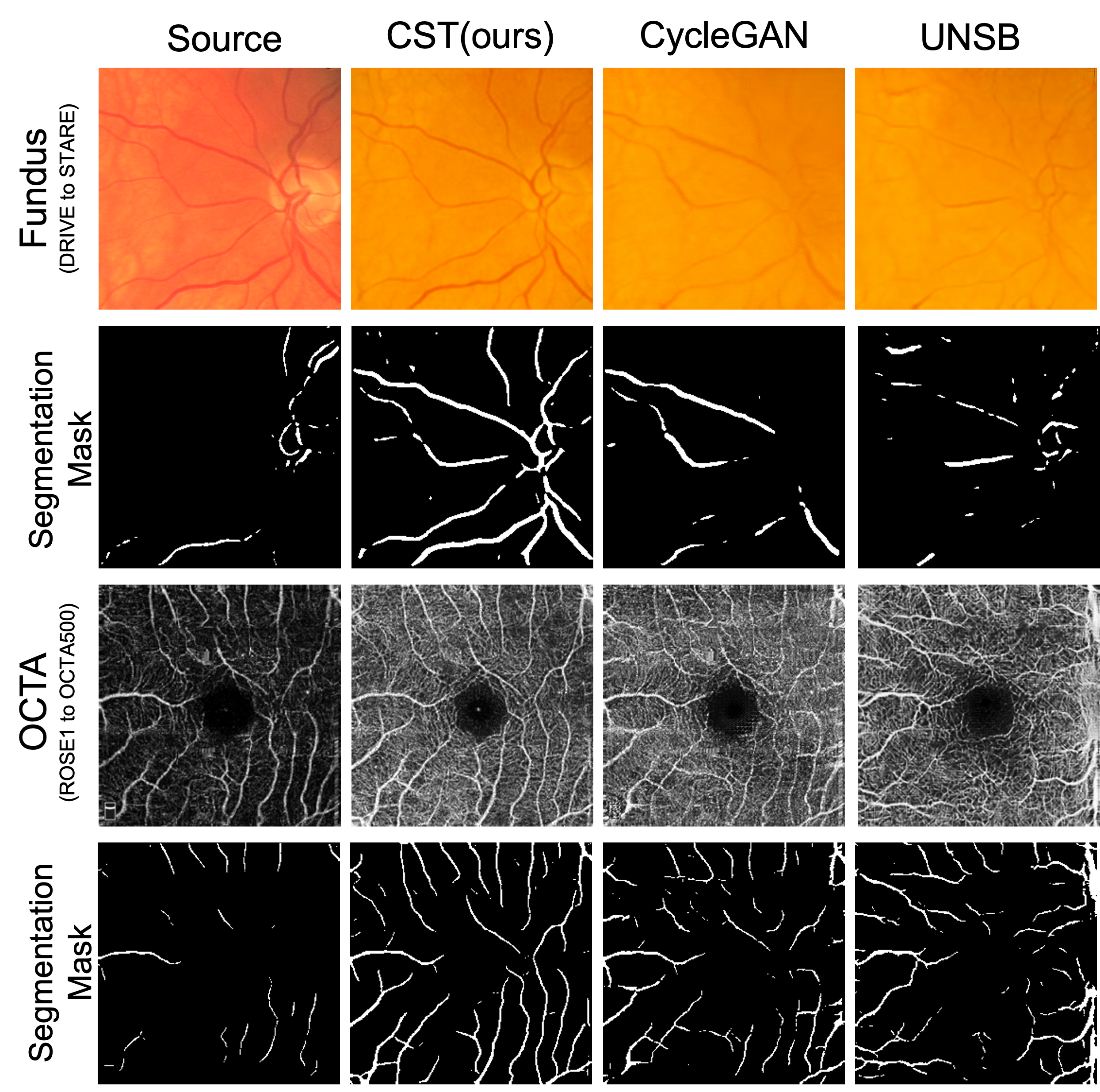}
  \caption{Comparison of domain translation results between the proposed method and other approaches, together with vessel segmentation mask visualizations. For fundus images, the source is from DRIVE~\cite{DRIVE}, and segmentation was carried out using a U-Net model trained on STARE~\cite{STARE}. For OCTA images, the source is from ROSE-1~\cite{ma2021rose:}, and segmentation was carried out using a U-Net model trained on OCTA-500~\cite{li2019octa}. The segmentation masks illustrate the structural fidelity of the translated images with respect to downstream vessel analysis.
  }
  \label{Seg_results}
\end{figure}

\subsection{Qualitative Results}

To evaluate the visual quality and structural preservation capabilities, we compare our method against four state-of-the-art unpaired image translation methods. Fig.~\ref{image_results} presents two representative translation results for the OCTA domain (OCTA500$\leftrightarrow$ROSE) where the first case is from a subject with diabetic retinopathy. While competing methods exhibit varying degrees of structural distortion—including vessel discontinuities, blurred boundaries, and loss of fine capillary networks—our approach can better preserve the intricate vascular architecture. Notably, the translated images maintain coherent vessel topology and morphological characteristics, which are critical for clinical interpretation and downstream diagnostic tasks.
The fundus photography translation results (DRIVE$\leftrightarrow$STARE) are illustrated in Fig.~\ref{fundus_results}. The regions demarcated by green boxes and arrowheads reveal that alternative methods frequently introduce artifacts such as fragmented vessels, spurious branching patterns, or thickness inconsistencies. In contrast, our method preserves the geometric continuity and anatomical fidelity of retinal vasculature across domain boundaries, maintaining vessels with high structural accuracy.
Fig.~\ref{xca} shows sample results    on X-ray coronary angiography (DCA1$\leftrightarrow$XCAD). The red arrowheads highlight regions where baseline methods fail to maintain vessel integrity, resulting in broken or distorted coronary arterial structures. These results demonstrate the model’s capability to generalize across diverse modalities while retaining curvilinear structure.

Fig.~\ref{Seg_results} illustrate the downstream segmentation tasks. Fundus images were translated from DRIVE to STARE and subsequently segmented using a vessel segmentation model trained on STARE. For OCTA, translation was performed from ROSE-1 to OCTA-500 with a 3$\times$3~mm$^2$ field of view. In both cases, the segmentation masks derived from the translated images indicate that our method achieves better vessel continuity and morphological preservation compared to other approaches. 

\subsection{Ablation Study}

\begin{table}[tb]
\caption{Ablation study evaluating the impact of different components.}
\centering
\begin{tabular}{l|cc|cc}
\toprule
\textbf{Configuration} & $\uparrow$\textbf{mDice} & $\uparrow$\textbf{mIoU} & $\downarrow$\textbf{LPIPS} & $\uparrow$\textbf{SSIM} \\
\midrule
Ours (UNSB) & \textbf{0.624} & \textbf{0.456} & \textbf{0.168} & \textbf{0.807} \\
w/o $\mathcal{L}_{\text{Cur}}$ & 0.556 & 0.389 & 0.246 & 0.788\\
w/o $\mathcal{L}_{\text{rot}}$ & 0.534 & 0.370 & 0.218 & 0.723\\
Baseline & 0.509 & 0.347 & 0.292 & 0.653\\
\bottomrule
\end{tabular}
\label{ablation_study}
\end{table}

\begin{figure}[!t]
  \centering
  \includegraphics[width=0.48\textwidth]{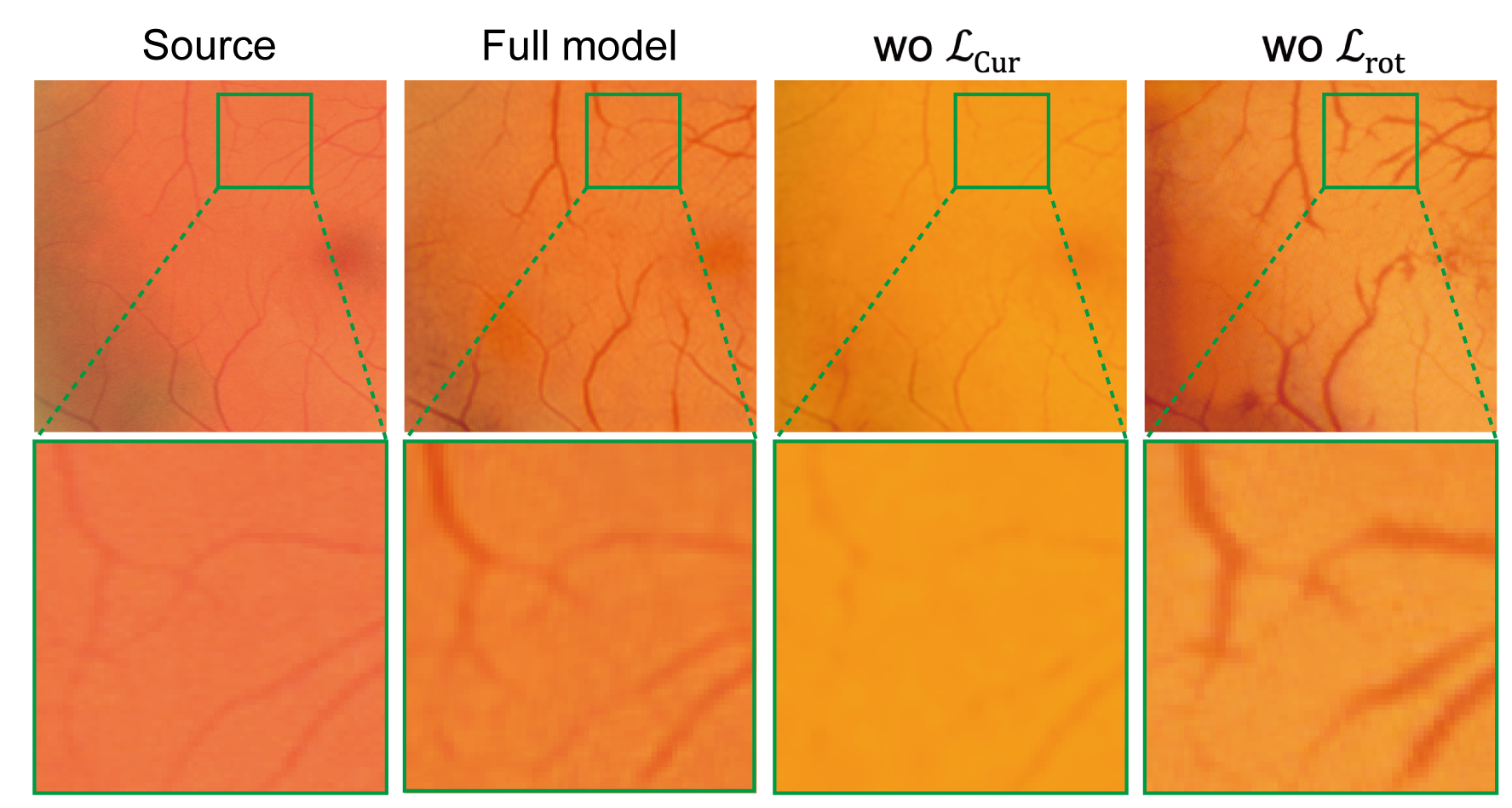}
  \caption{Ablation study on domain translation results. The figure compares the outputs of the full model with those obtained by removing individual components, illustrating the contribution of each module to the overall performance.
  }
  \label{ablation}
\end{figure}
To evalaute the effectiveness of the proposed components, a series of ablation experiments was conducted. We use the color fundus translation from   DRIVE  to STARE as an example. The quantitative results are reported in Table~\ref{ablation_study}. 

Removing the curvilinear structure loss leads to  relative  declines of 10.9\% and 14.7\% in mDice and mIoU respectively. Excluding the rotation consistency loss results in relative declines of 14.4\% and 18.9\% respectively. As expected, removing both components yields the lowest overall performance.

\begin{table}[t]
\centering
\caption{Comparison of selecting different extractors on the image translation task from DRIVE to STARE. We use Ours (UNSB) as the translation framework and replace only the CEM. 
The metrics are computed by applying a segmentation model trained on the target domain to the translated images.}
\label{whyucs}
\resizebox{\columnwidth}{!}{
\begin{tabular}{lccc}
\toprule
\textbf{Extractor} & \textbf{mDice} $\uparrow$ & \textbf{mIoU} $\uparrow$ & \textbf{clDice} $\uparrow$ \\
\midrule
SAM \cite{SAM} & 0.608$\pm$0.066 & 0.440$\pm$0.067 & 0.577$\pm$0.081 \\
SAM-Med2D \cite{cheng2023sammed2d} & 0.607$\pm$0.073 & 0.439$\pm$0.073 & 0.571$\pm$0.083 \\
UCS \cite{UCS} (Ours) & \textbf{0.624$\pm$0.063} & \textbf{0.456$\pm$0.065} & \textbf{0.608$\pm$0.075} \\
\bottomrule
\end{tabular}
}
\end{table}

To validate our selection of UCS \cite{UCS} as the CEM, we compare it against two popular foundation models: SAM~\cite{SAM} and SAM-Med2D~\cite{cheng2023sammed2d}. As shown in Table~\ref{whyucs}, UCS achieves superior performance compared to SAM  and SAM-Med2D. This demonstrates that CEM adopted from UCS is more effective at capturing fine-grained vessel topology and preserving structural continuity during domain translation, whereas general-purpose segmentation models failed to identify thin, elongated structures characteristic of vessels.

\subsection{Generalization to Other Data}
\begin{table}[t]
\caption{
Segmentation results (mean $\pm$ std). The results are shown for translated images from GAPs384 to CFD. The segmentation model was trained on the CFD.}
\centering
\setlength{\tabcolsep}{2.5mm}
\begin{tabular}{l|ccc}
\toprule
\textbf{Method} & $\uparrow$mDice & $\uparrow$mIoU & $\uparrow$clDice \\
\midrule
No Translation & 0.138$\pm$0.187 & 0.087$\pm$0.123 & 0.164$\pm$0.220 \\
CycleGAN \cite{CycleGAN2017} & 0.254$\pm$0.160 & 0.156$\pm$0.111 & 0.312$\pm$0.193 \\
CUT \cite{park2020cut} & 0.135$\pm$0.105 & 0.076$\pm$0.063 & 0.165$\pm$0.128 \\
SynDiff \cite{ozbey_dalmaz_syndiff_2024} & 0.042$\pm$0.075 & 0.023$\pm$0.045 & 0.042$\pm$0.080 \\
UNSB \cite{UNSB} & 0.143$\pm$0.136 & 0.084$\pm$0.087 & 0.162$\pm$0.158 \\
\midrule
\textbf{Ours (UNSB)} & \underline{0.291}$\pm$\underline{0.170} & \underline{0.182}$\pm$\underline{0.119} & \underline{0.327}$\pm$\underline{0.198} \\
\textbf{Ours (CycleGAN)} & \textbf{0.412}$\pm$\textbf{0.188} & \textbf{0.276}$\pm$\textbf{0.144} & \textbf{0.485}$\pm$\textbf{0.219} \\
\bottomrule
\end{tabular}
\label{tab:seg_pavement}
\end{table}

\begin{figure}[!t]
  \centering
  \includegraphics[width=0.48\textwidth]{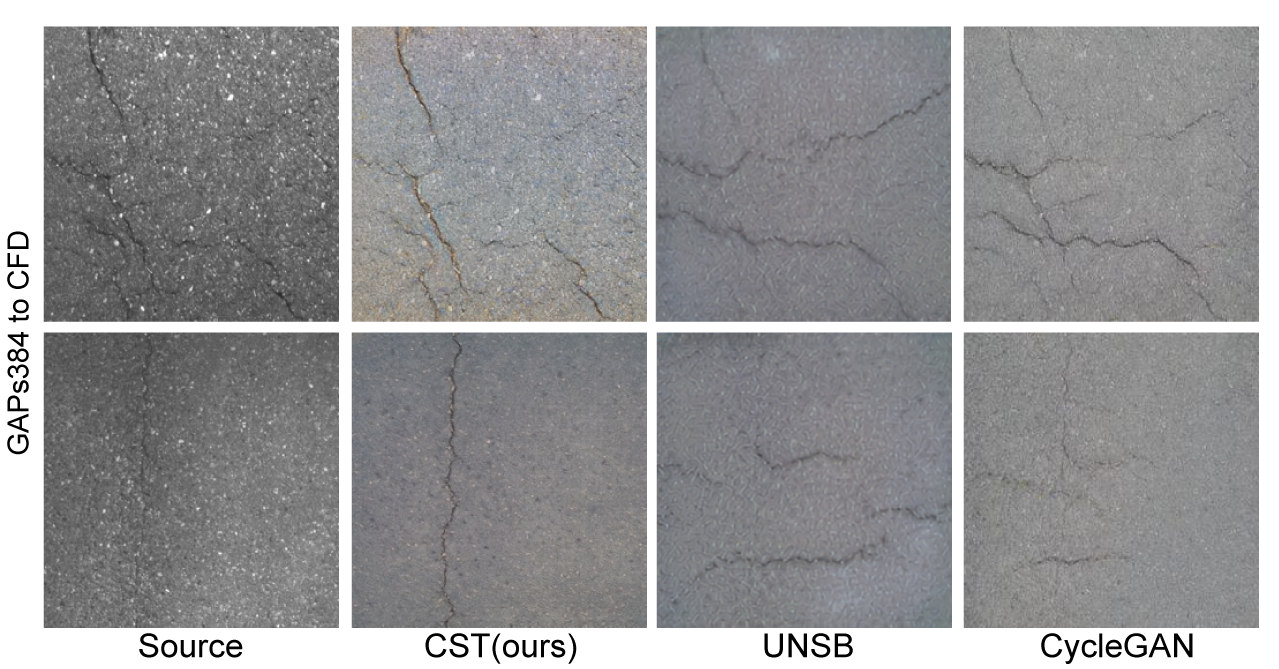}
  \caption{Qualitative results on the pavement task. Our method generalizes beyond medical data to natural images, effectively preserving fine crack structures that CycleGAN and UNSB fail to maintain.
  }
  \label{fig:pavement}
\end{figure}

Although primarily designed for medical image translation, our proposed method exhibits considerable potential for extension to natural image processing applications, including automated pavement crack detection. This transferability is grounded in several fundamental algorithmic and structural correspondences between the two domains.

Experimental validation was conducted using the GAPs384~\cite{GAPs384} and CFD~\cite{CFD} pavement crack datasets, which present distinct challenges including nighttime imaging conditions and diverse surface textures spanning both concrete and asphalt materials. The results are summarized in Table~\ref{tab:seg_pavement}.  Fig.~\ref{fig:pavement} shows two example for visual illustration. The results show that our method preserves cracks well during translation, as the translated images exhibit strong cross-domain generalization capabilities in downstream segmentation tasks.
This demonstrates the generalizability of our approach and establishes foundations for broader deployment in image translation  applications requiring detailed curvilinear structural analysis under challenging imaging conditions.

\section{Conclusion}
In this work, we present the CST, the Curvilinear Structure-preserving Translation framework designed for unpaired image translation across domains. A central challenge in this setting lies in preserving fine curvilinear structures, such as blood vessels in medical imagery, which are vital for downstream tasks like segmentation and medical analysis. CST explicitly enforces structural integrity during translation and is evaluated on three representative modalities: OCTA, fundus, and XCA. Experimental results demonstrate that CST consistently surpasses existing methods in both perceptual quality and structural fidelity, leading to superior translation performance and improved downstream segmentation. These findings underscore the robustness and generalizability of CST for translating images with curvilinear structures across diverse domains. 
Beyond these results, CST provides a general paradigm for translating curvilinear structure-rich biomedical images, where preserving topology and connectivity is critical. Its structure-aware formulation can be extended to other modalities such as neural imaging, or integrated with generative priors for data augmentation and model pretraining. This work thus lays a foundation for more reliable and anatomically consistent image translation in medical and scientific applications. 

\appendices

\bibliographystyle{IEEEtran}
\bibliography{reference}

\end{document}